\documentclass[sigconf]{acmart}

\usepackage{amsmath}

\usepackage{amssymb}

\usepackage{multirow}
\usepackage{animate}
\usepackage{makecell}
\usepackage{enumitem}
\usepackage{marvosym}

\renewcommand\footnotetextcopyrightpermission[1]{}

\AtBeginDocument{
  \providecommand\BibTeX{{
    \normalfont B\kern-0.5em{\scshape i\kern-0.25em b}\kern-0.8em\TeX}}}

\setcopyright{acmcopyright}
\copyrightyear{2018}
\acmYear{2018}
\acmDOI{XXXXXXX.XXXXXXX}

\acmConference[Conference acronym 'XX]{Make sure to enter the correct
  conference title from your rights confirmation emai}{June 03--05,
  2018}{Woodstock, NY}
\acmPrice{15.00}
\acmISBN{978-1-4503-XXXX-X/18/06}

\begin{document}

\title{ROMA: Cross-Domain Region Similarity Matching for Unpaired\\
Nighttime Infrared to Daytime Visible Video Translation}

\author{Zhenjie Yu}
\affiliation{%
  \institution{Beijing Institute of Technology}
  \state{Beijing}
  \country{China}
}
\email{zjyu@bit.edu.cn}

\author{Kai Chen}
\affiliation{%
  \institution{Yantai IRay Technologies Lt. Co.}
  \state{Shandong}
  \country{China}
}
\email{kai.chen@iraytek.com}

\author{Shuang Li}
\affiliation{%
  \institution{Beijing Institute of Technology}
  \state{Beijing}
  \country{China}
}
\email{shuangli@bit.edu.cn}
\authornote{Corresponding author: Shuang Li.}

\author{Bingfeng Han}
\affiliation{%
  \institution{Beijing Institute of Technology}
  \state{Beijing}
  \country{China}
}
\email{bfhan@bit.edu.cn}

\author{Chi Harold Liu}
\affiliation{%
  \institution{Beijing Institute of Technology}
  \state{Beijing}
  \country{China}
}
\email{liuchi02@gmail.com}

\author{Shuigen Wang}
\affiliation{%
  \institution{Yantai IRay Technologies Lt. Co.}
  \state{Shandong}
  \country{China}
}
\email{shuigen.wang@iraytek.com}

\begin{abstract}
Infrared cameras are often utilized to enhance the night vision since the visible light cameras exhibit inferior efficacy without sufficient illumination. However, infrared data possesses inadequate color contrast and representation ability attributed to its intrinsic heat-related imaging principle. This makes it arduous to capture and analyze information for human beings, meanwhile hindering its application. Although, the domain gaps between unpaired nighttime infrared and daytime visible videos are even huger than paired ones that captured at the same time, establishing an effective translation mapping will greatly contribute to various fields. In this case, the structural knowledge within nighttime infrared videos and semantic information contained in the translated daytime visible pairs could be utilized simultaneously. To this end, we propose a tailored framework \emph{ROMA} that couples with our introduced c\textbf{R}oss-domain regi\textbf{O}n si\textbf{M}ilarity m\textbf{A}tching technique for bridging the huge gaps. To be specific, ROMA could efficiently translate the unpaired nighttime infrared videos into fine-grained daytime visible ones, meanwhile maintain the spatiotemporal consistency via matching the cross-domain region similarity. Furthermore, we design a multiscale region-wise discriminator to distinguish the details from synthesized visible results and real references. Extensive experiments and evaluations for specific applications indicate ROMA outperforms the state-of-the-art methods.  Moreover, we provide a new and challenging dataset encouraging further research for unpaired nighttime infrared and daytime visible video translation, named \emph{InfraredCity}. In particular, it consists of 9 long video clips including City, Highway and Monitor scenarios. All clips could be split into $603,142$ frames in total, which are $20$ times larger than the recently released daytime infrared-to-visible dataset IRVI.
  
\end{abstract}


\begin{teaserfigure}
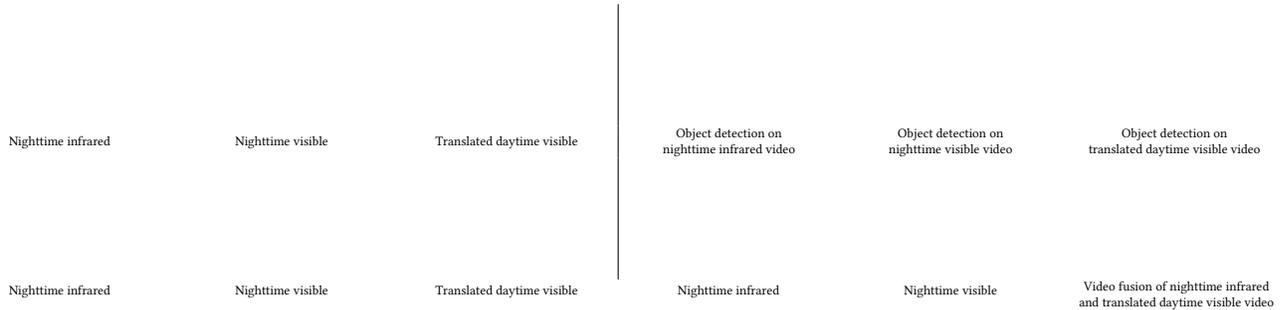

  \centering
  \tiny
  \begin{tabular}{m{26mm}<{\centering} m{26mm} <{\centering} m{26mm} <{\centering} m{26mm} <{\centering} m{26mm} <{\centering} m{26mm} < {\centering}}
    \multicolumn{3}{c}{\animategraphics[autoplay,loop,width=0.48\textwidth]{30}{images/traffic/0000}{00}{60}}
    &\multicolumn{3}{|c}{\animategraphics[autoplay,loop,width=0.48\textwidth]{15}{images/detection/0000}{00}{60}}
    \\
    \makecell[c]{Nighttime infrared}&\makecell[c]{Nighttime visible}&\makecell[c]{Translated daytime visible}  & \multicolumn{1}{|c}{\makecell[c]{Object detection on\\nighttime infrared video}} & \multicolumn{1}{c}{\makecell[c]{Object detection on\\nighttime visible video}} &\multicolumn{1}{c}{\makecell[c]{Object detection on\\translated daytime visible video}}
    \\ 
    \multicolumn{3}{c}{\animategraphics[autoplay,loop,width=0.48\textwidth]{30}{images/monitor/0000}{05}{65}}
    &\multicolumn{3}{|c}{\animategraphics[autoplay,loop,width=0.48\textwidth]{15}{images/fuse/0000}{00}{60}}
    \\
    \makecell[c]{Nighttime infrared}&\makecell[c]{Nighttime visible}&\makecell[c]{Translated daytime visible}  & \makecell[c]{Nighttime infrared} & \makecell[c]{Nighttime visible} &\makecell[c]{Video fusion of nighttime infrared\\and translated daytime visible video}
  \end{tabular}
  \caption{Left: Display of the nighttime infrared, nighttime visible and translated daytime visible videos via ROMA for Highway and Monitor scenarios in InfraredCity, respectively. 
  Right: Display of applications on object detection and video fusion for ROMA-translated results. Notably, the ROMA-translated daytime videos achieve superior detection performance compared with the corresponding nighttime infrared and visible videos, and the video fusion results between infrared and ROMA-translated videos are much sharper than the counterparts. \emph{The animated videos are best viewed via Adobe Acrobat, please zoom in for details.}}
  \label{fig:start}
\end{teaserfigure}

\maketitle

\section{Introduction}
In real-world multimedia applications, visible light cameras are often leveraged to assist improving visual effects for various scenarios \cite{2018fusion}. Unfortunately, their adaptability becomes even worse than the human biological vision system under extreme conditions, e.g., dark night or light exposure as shown in Fig.~\ref{fig:start}. In such cases, infrared sensors could take the place of visible cameras as an auxiliary imaging systems. Its heat-related imaging principle can stably provide visual signals with sufficient spatial or structural descriptions, however, the lack of detailed semantic information can not well satisfy the conventional cognition of this colorful world \cite{zhou2016fusion}. This makes the raw infrared data undesirable for direct utilization in practical tasks, such as autonomous driving or monitoring. Even so, infrared sensors are indispensable in real-world applications, especially for the dark night situation. Therefore, it is worthy to bridge the modality gaps between the infrared and visible data.

Plenty of works~\cite{fusion, zhou2016fusion, BavirisettiXZDL19, lowRankFusion} have thoroughly studied the data fusion for infrared and visible frames, while their results often visually retain the appearance of grayscales, which are still not distinct compared with visible ones.
Besides, other methods~\cite{colorization2016, 2006-manga, 2016FACE, TCSVT} try to transform infrared images to visible ones through different color mapping functions. However, they usually require complex manual interventions, which impose limitations in practical applications.
With the development of deep learning based generative models, image-to-image translation methods~\cite{cycleGAN, CUT, PCSGAN, FLSeSim} have attracted appreciable attentions. They seek to achieve high-quality performance through powerful generative adversarial training techniques~\cite{GAN}. However, the huge domain gaps between infrared and visible images make these methods incompetent in precisely preserving the proper infrared structure information and abundant visible semantic details. Moreover, nowadays most real-world applications provide feedback in the form of video signals. These image-level methods will lose their applicability due to lack of consideration for video temporal consistency.

To this end, video-to-video translation methods have recently taken a further step on the basis of their counterparts. For instance, \cite{fewshot-vid2vid,vid2vid} synthesize target videos and predict future frames greatly rely on labeled pair data. However, like most video translation tasks, the nighttime infrared and daytime visible videos have no pixel-to-pixel aligned training data. Manual labeling is not only time-consuming and expensive, but also prone to introduce errors.
Thus, some unpaired video-to-video translation methods~\cite{uvid2vid, RecycleGAN, MocycleGAN, I2V} loosen the requirements for paired data. In particular, \cite{I2V} is a framework that adapts for daytime infrared and visible video translation. The huger domain gap between nighttime infrared and daytime visible data will inevitably diminish their translated performance, since the critical semantic structural correspondence across the two domains are largely overlooked. Moreover, their feature-level losses have integrated the domain-invariant content and domain-specific appearance together, and these constrains may confuse these attributes in the generative process. Hence, the structural consistency and semantic details will be destroyed in the translated results, especially for tremendous domain discrepancy.

To tackle these challengings, we propose a novel one-sided end-to-end framework \emph{ROMA} that couples with our introduced cross-domain region similarity matching technique (\emph{referred to as cross-similarity in the rest of this paper}) to bridge the huge gap between nighttime infrared and daytime visible videos.  
Specifically, we focus on the domain-invariant structural information by optimizing three forms of cross-similarity maps to generate fine-grained translating results and keep temporal consistency. Besides, we propose a multiscale region-wise discriminator to enhance the detailed domain-specific style information.

\begin{figure}[!t]
    \centering
    \includegraphics[width=8.3cm]{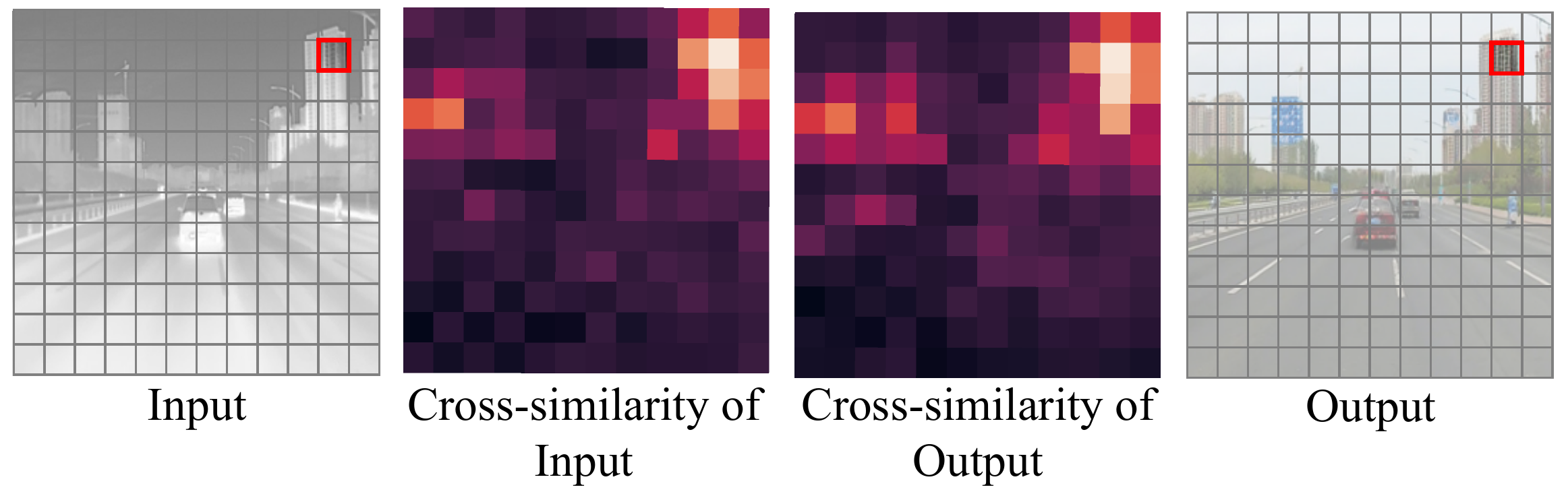}
    \caption{Example of our learned input and synthesized output cross-similarity map, which is calculated as the token similarity between the boxed region (key) in input/output and all the regions (queries) in output/input. The matching of cross-domain region similarity maps is imposed to enhance the content and structure correspondence between input and output, while getting rid of the domain style effects.}
    \vspace{-0.22in}
    \label{fig:intro}
\end{figure}

As shown in Fig.~\ref{fig:intro}, the cross-similarity map is calculated between input and synthesized output on the basis of a pre-trained ViT \cite{ViT}. We first split the frames into several regions, and each region of the input could obtain similarity maps interacting with all regions of the output. Meanwhile, the similarity maps of the output regions could also obtain by interacting with the input regions. Notably, all the cross-similarity maps are calculated across domains, which could make the generative process focus on learning the content or structural correspondence between real and synthesized frames, while getting rid of the negative effects of different styles. Then we match the cross-domain region similarity maps calculated at the same location. This process is called \emph{global structural cross-similarity consistency}, since it keeps the domain-invariant structural information to be the same under different appearances. Moreover, we take a step further to apply this pattern within local regions for fine-grained structure preservation, which is called \emph{local structural cross-similarity consistency}. Since the cross-similarity map matching cuts off the influence of domain-specific style information, the similarity maps can also measure structural variants caused by scene movements. Thus, we optimize \emph{temporal cross-similarity consistency} by minimize the distance of similarity maps that calculated from time $t$ input, time $t+1$ output and time $t$ output, time $t+1$ input.

As for domain-specific style enhancement, we propose a \emph{multiscale region-wise discriminator}. We first extract the token embeddings of real references and synthesized results via ViT, then reshape and refine them into different scales. At last, the MLP layers are applied to distinguish real or fake depending on the concatenation of multiple representations.

Additionally and importantly, the nighttime infrared to daytime visible video translation is rarely studied due to the lack of a high-quality relevant dataset. In this paper, we offer a new and challenging dataset named \emph{InfraredCity}, which consists of 9 long video clips including city, highway, and monitoring scenarios. All clips could be split into $603,142$ frames in total. After manual selection, we additionally provide \emph{InfraredCity-Lite} for research. In summary, our contributions are listed as below:

\begin{itemize}[leftmargin=10pt]
    \item A simple yet effective cross-domain region similarity matching technique is proposed, which could fully utilize the structure knowledge of nighttime infrared data, and enhance the structure correspondence between input and output, facilitating generating authentic and fluent daytime visible videos.
    \item We propose an efficient one-sided end-to-end framework ROMA, which performs \emph{cross-similarity matching} and is coupled with our introduced \emph{multiscale region-wise discriminator}. Besides, ROMA achieves superior performance compared to other state-of-the-art baselines on several datasets.
    \item The translated videos via ROMA could be further applied to real-world applications, such as object detection and video fusion. The promising results validate the effectiveness of ROMA for night vision scenarios.
    \item We provide new datasets for nighttime infrared to daytime visible video translation, i.e., \emph{InfraredCity} and \emph{InfraredCity-Lite}, encouraging further research on this area. 
\end{itemize}

\section{Related Work}

\textbf{Infrared-to-Visible Translation.} 
Since infrared-to-visible translation is an attractive strategy to enhance nighttime vision perception, there are continuing studies on it. Generally, the infrared vision technique is often used for the context enhancement in nighttime vision by fusing it with the visible data \cite{zhou2016fusion, fusion, fusion}. Unfortunately, the nighttime visible image is often terrible on account of low-light conditions, which delivers limited knowledge to infrared data. Also, the gray fusion results are undesirable for human beings. Thus, \cite{2006-manga, TCSVT, 2002gray, 2003color} regard infrared images as gray ones and attempt to generate visible images via colorization approaches. Similarly, \cite{2007color, IRcolor2016, DCGANInfra, LCInfra} utilize the GAN \cite{GAN} module and attempt to affine the single-channel infrared image into the three-channel RGB result via colorization manners. Although these methods could generate colorful results, they prone to distort details without additional structural constraints. Moreover, these image-level methods impose limitations on the infrared-to-visible video translation task due to lack of consideration for temporal consistency.

\textbf{Image-to-Image and Video-to-Video Translation.}
Image-to-image translation intends to learn a mapping from the source domain to the target domain.
Pix2pix \cite{pix2pix} explores the possibility of applying the deep network on image translation via the GAN framework \cite{GAN} with paired datasets. Furthermore, to relax the requirements of paired datasets, CycleGAN \cite{cycleGAN} introduces the cycle consistency, which maintains the content during training. However, it requires auxiliary generators and discriminators for the reverse mapping, leading to more computation cost. To avoid this, \cite{onesidemaping, GcGAN} adopt a one-sided framework and propose implicit structural consistency to replace cycle consistency. \cite{perceptual, Contextual, 2017Photographic} propose structural consistency under the high-level semantic information. CUT \cite{CUT} and its following algorithm \cite{PWSR} seek to maximize the mutual information between the two domains. 
Although these approaches guarantee spatial consistency in image-to-image translation tasks, they can not be directly applied to the video translation tasks due to the lack of consideration for temporal coherence.

To make up for the deficiency, \cite{fewshot-vid2vid,vid2vid} translate videos with hand-designed temporal consistency on the basis of paired video datasets. However, it is almost impossible to collect the pixel-to-pixel paired videos for the nighttime infrared and daytime visible translation tasks. Thus, \cite{uvid2vid} explores unpaired video-to-video translation and proposes 3D Convolution as a pioneer. RecycleGAN \cite{RecycleGAN} utilizes cycle loss and recurrent loss to keep the consistency of spatial and temporal information. Similarly, mocycle-GAN \cite{MocycleGAN} utilizes cycle loss for structural consistency and motion consistency loss based on optical flow to maintain the temporal coherence. Especially, I2V-GAN \cite{I2V} is a tailored approach for daytime infrared-to-visible translation based on the Recycle-GAN framework, which additionally proposes perceptual cyclic losses and similarity losses to enhance the spatiotemporal consistency. Although I2V-GAN achieves acceptable results on infrared-to-visible translation, huger domain gaps between nighttime infrared and daytime visible data hinder its efficacy. Besides, these approaches perform their constraints mainly on image level and feature level which will inevitably conflate video content and style information together in the optimization process due to the huge difference between two domains. Therefore, the intrinsic structural correlation between each region in the infrared videos will be entangled with the domain style, leading to generating blurred and unpleasant visible videos.

\section{The Proposed Method}

\begin{figure}[!t]
    \centering
    \includegraphics[width=8.3cm]{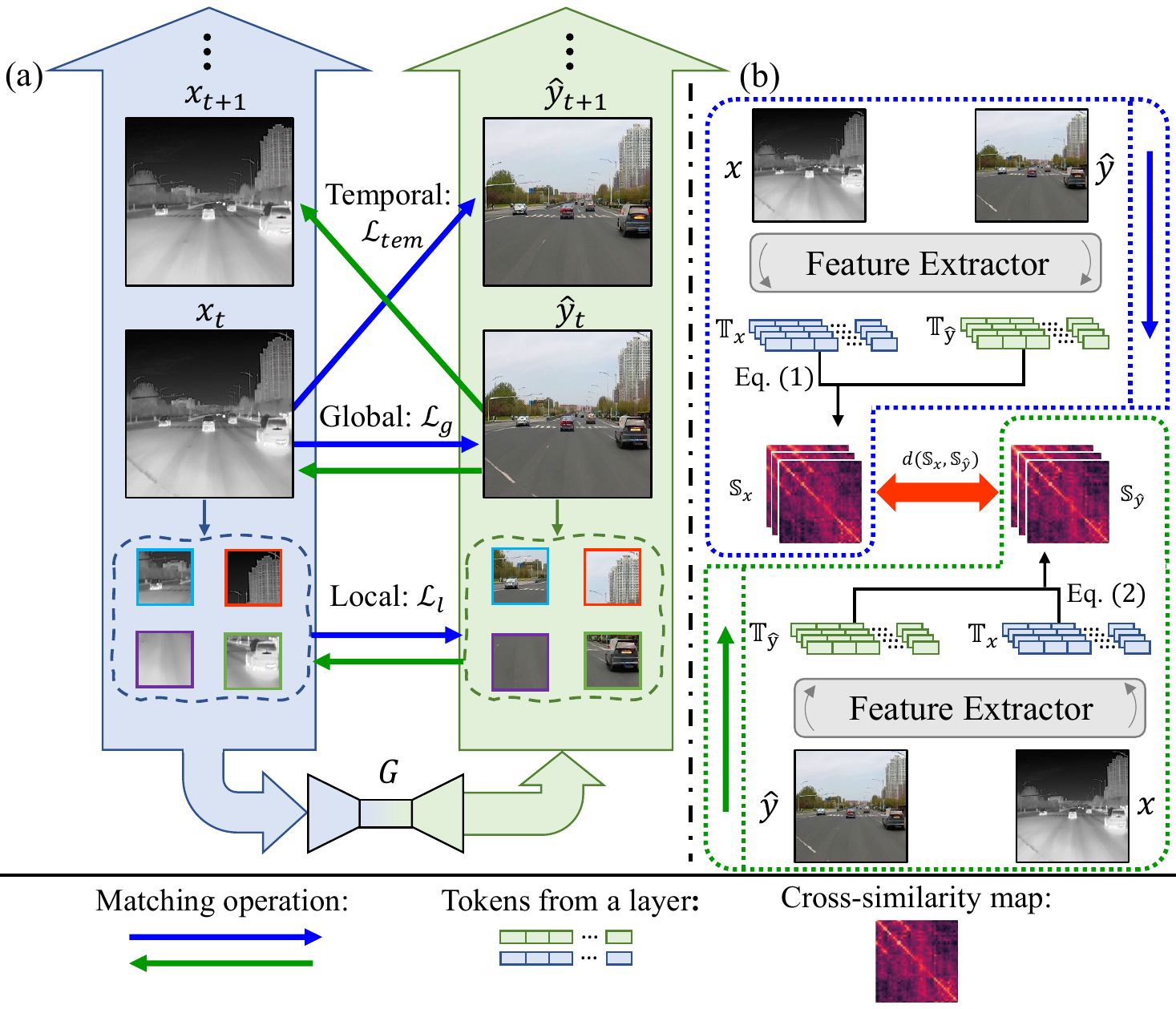}
    \vspace{-1em}
    \caption{(a) Three forms of \textit{cross-similarity} constraints for nighttime infrared to daytime visible translation in ROMA. (b) The calculation procedure of cross-similarity matching.}
    \label{fig:cross-correlation}
    \vspace{-0.22in}
\end{figure}

For an infrared video clip $\mathbb{X} = \{x_1, x_2,...,x_N~|~x\in\mathbb{R}^{H\times W\times C}\}$ and a visible video clip $\mathbb{Y}=\{y_1, y_2,...,y_{N'}~|~y\in\mathbb{R}^{H\times W\times C}\}$, we aim to guide the generator $G$ to transfer $\mathbb{X}$ into the target-style clip $\hat{\mathbb{Y}} = \{\hat{y}_1, \hat{y}_2,...,\hat{y}_N~|~\hat{y}\in \mathbb{R}^{H\times W\times C}\}$. In particular, the translated results $\hat{\mathbb{Y}}$ should maintains the spatiotemporal consistency with $\mathbb{X}$, but converts the appearance appropriately as real visible video ${\mathbb{Y}}$. Notably, ${\mathbb{Y}}$ and $\hat{\mathbb{Y}}$ only share the same style information within their different scenes. This translation process is denoted as $\hat{\mathbb{Y}}=G(\mathbb{X})$.

We begin this section by introducing our proposed \textit{cross-domain region similarity matching} technique. Specifically, we propose three types of constraints from both spatial and temporal perspectives for video translation, which concentrate on the domain-invariant information and cut off the negative influence of domain-specific information.
Then, we introduce a \textit{multiscale region-wise discriminator}, which is applied to distinguish the details of domain-specific information between synthesized results and real references.

\subsection{Cross-domain Region Similarity Matching}\label{method3.1}

We take an input frame $x \in \mathbb{X}$ and the corresponding synthesized output $\hat{y} \in \hat{\mathbb{Y}}$ as an example to illustrate. 
As shown in Fig.~\ref{fig:intro}, we first divide $x$ and $\hat{y}$ into non-overlapping regions. Then the cross-similarity map of $i$-th \textbf{s}ource \textbf{r}egion $sr_i$ in $x$ can be calculated associate to all regions of $\hat{y}$. Similarly, the cross-similarity map of $i$-th \textbf{t}arget \textbf{r}egion $tr_i$ is obtained by interacting with all the source regions. We formulate this process for cross-similarity maps calculation as following:
\begin{align}
    S_{sr_i} &= u_i \cdot v_*^\top, \\
    S_{tr_i} &= v_i \cdot u_*^\top,
\label{eq1}
\end{align}
where $u_i$, $v_i \in \mathbb{R}^{1\times d}$ are d-dimensional token embeddings for region $sr_i$ and region $tr_i$, $u_*^\top$, $v_*^\top \in \mathbb{R}^{d\times N_r}$ denote the transposed token embeddings of all $N_r$ non-overlapping regions in $x$ and $\hat{y}$, respectively. Thus, $S_{sr_i}$ stands for the cross-similarity map of region $sr_i$ in $x$, and $S_{tr_i}$ stands for the cross-similarity map of region $tr_i$ in $\hat{y}$. 
After we obtain the cross-similarity collections $S_x = [S_{sr_1}, S_{sr_2},...,S_{sr_{N_r}}]$ and $S_{\hat{y}} = [S_{tr_1}, S_{tr_2},...,S_{tr_{N_r}}]$, 
we perform the similarity matching by minimizing their distance within multiple layers of the ViT. Specifically, by picking several representations from different layers, we get $\mathbb{S}_x = [S_x^{l_1}, S_x^{l_2},...,S_x^{l_{N_l}}]$ and $\mathbb{S}_{\hat{y}} = [S_{\hat{y}}^{l_1}, S_{\hat{y}}^{l_2},...,S_{\hat{y}}^{l_{N_l}}]$, which represent different views of $x$ and $\hat{y}$. Finally, the objective of cross-similarity matching between $x$ and $\hat{y}$ is defined as:
\begin{align}
    \mathcal{L}&=\frac{1}{N_l}d(\mathbb{S}_x, \mathbb{S}_{\hat{y}})=\frac{1}{N_l}\sum^{N_l}_{j=1}d(S_x^j, S_{\hat{y}}^j),
\end{align}
where $d(\cdot)$ is the cosine distance function which performs better than $L1$ and $L2$ according to numerous experimental results.
Based on this cross-domain region similarity, we design three forms of constraints for spatiotemporal consistency shown in Fig.~\ref{fig:cross-correlation}.

\textbf{Global Cross-Similarity for Structural Consistency.} Given input frames $\mathbb{X}=\{x_1, x_2,...,x_N\}$ from source domain and their corresponding synthesized target frames $\hat{\mathbb{Y}}=\{\hat{y}_1, \hat{y}_2,...,\hat{y}_N\}$, the global cross-similarity matching performs as:
\begin{align}
    \mathcal{L}_{g}(G) = \frac{1}{N_{l}}\sum^{N}_{t=1}d(\mathbb{S}_{x}(t),\mathbb{S}_{\hat{y}}(t)),
\end{align}
where $\mathbb{S}_{x}(t)$ and $\mathbb{S}_{\hat{y}}(t)$ are the collections of cross-similarity maps of the $t$-th frames in $\mathbb{X}$ and $\hat{\mathbb{Y}}$ that obtained from different $l$ layers.

\begin{figure}[!t]
    \centering
    \includegraphics[width=8.5cm]{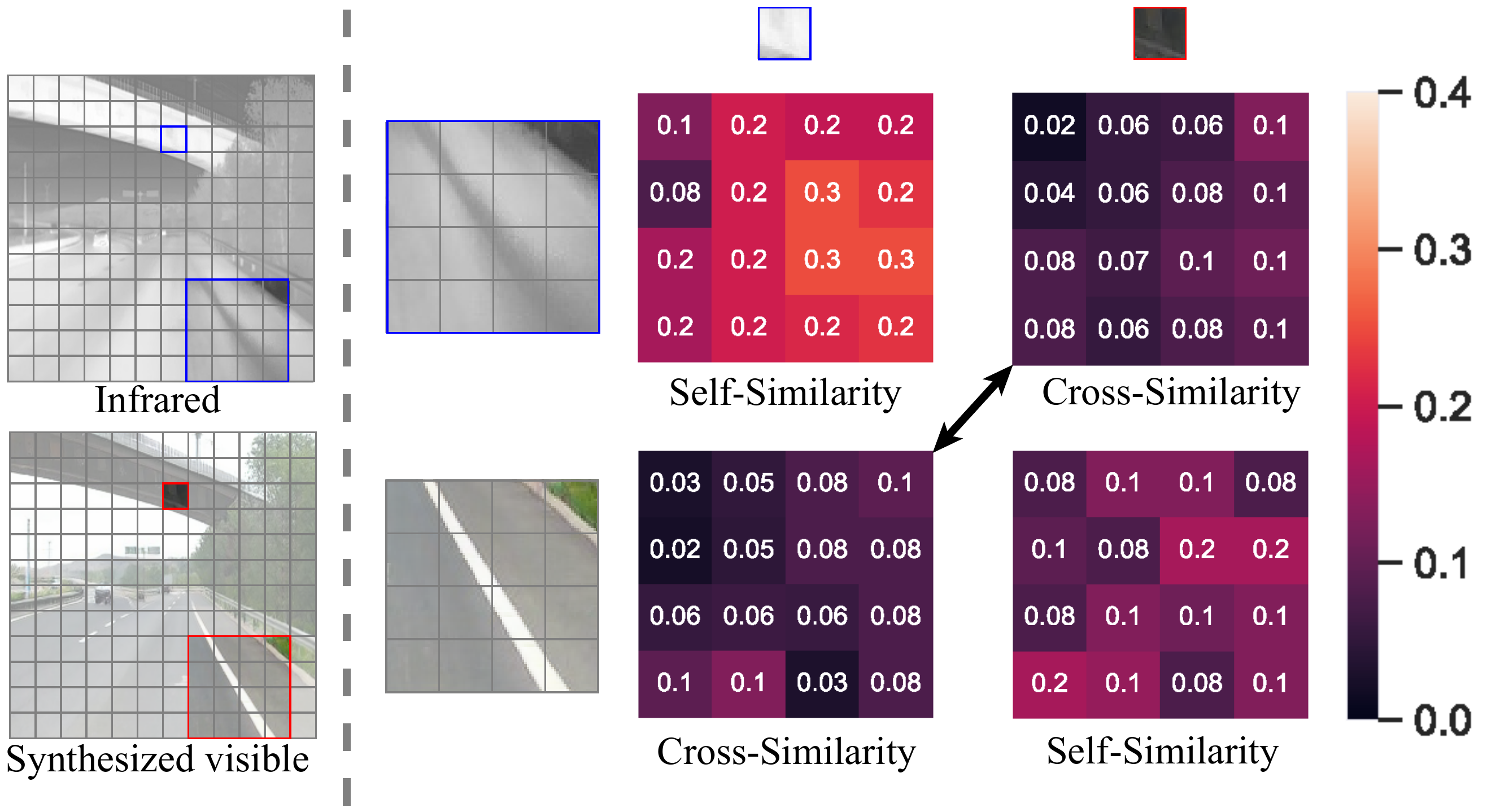}
    \caption{The visualization of cosine similarity calculated between irrelevant regions. It is obvious that cross-similarity can avoid the negative influences of domain-specific styles.}
    \label{fig:similarity}
    \vspace{-0.2in}
\end{figure}

\textbf{Local Cross-Similarity for Structural Consistency.}
Moreover, we propose a local optimization strategy to further improve the structural details for fine-grained translation results. For instance, as shown in Fig.~\ref{fig:cross-correlation}, we first randomly pick $N_a$ areas from the input frame $x_t$. The size of an area is larger than a region but smaller than a frame. Then, the areas of $\hat{y}_t$ are automatically obtained with the same locations as their corresponding areas in $x_t$. After that, the local cross-similarity matching is performed within each areas pair as:
\begin{align}
    \mathcal{L}_{l}(G) = \frac{1}{N_{l}}\frac{1}{N_a}\sum^{N}_{t=1}\sum^{N_a}_{i=1}d(\mathbb{S}^i_{x}(t),\mathbb{S}^i_{\hat{y}}(t)),
\end{align}
where $\mathbb{S}^i_{x}(t)$ and $\mathbb{S}^i_{\hat{y}}(t)$ are the collections of cross-similarity maps of the $i$-th areas in $t$-th frames from $\mathbb{X}$ and $\hat{\mathbb{Y}}$ from different $l$ layers.

\textbf{Cross-Similarity for Temporal Consistency.} 
As shown in Fig.~\ref{fig:intro} and Fig.~\ref{fig:similarity}, cross-similarity can cut off the negative influences of domain-specific styles and focus on the domain-invariant structure. Thus, the similarity maps can apply to measure the structural variants caused by scene movements and further optimize the temporal consistency for video translation.
To be specific, we utilize $u_{i}(t)$ and $v_{i}(t)$ to represent the $i$-th regions of $x_t$ and $\hat{y}_t$ at time $t$. $u_{*}(t+1)$ and $v_{*}(t+1)$ denote the token embeddings for all regions of $x_{t+1}$ and $\hat{y}_{t+1}$. The temporal cross-similarity is:
\begin{align}
    S_{sr_{i}}(t,t+1) = u_{i}(t)\cdot v_{*}(t+1)^\top, \\
    S_{tr_{i}}(t,t+1) = v_{i}(t)\cdot u_{*}(t+1)^\top.
\end{align}
We formulate the multilayer temporal cross-similarity matching as:
\begin{align}
    \mathcal{L}_{tem}(G) = \frac{1}{N_l}\sum^{N-1}_{t=1}d(\mathbb{S}_{x}(t,t+1), \mathbb{S}_{\hat{y}}(t,t+1)),
\end{align}
where $\mathbb{S}_{x}(t,t+1)$ is the collection $\{S_{sr_{i}}(t,t+1)~|~i=1,...,N_{r}\}$, and $\mathbb{S}_{\hat{y}}(t,t+1)$  is the collection $\{S_{tr_{i}}(t,t+1)~|~i=1,...,N_{r}\}$. Moreover, this constraint can be extended to a generic version related to length of video fragment in each training step:
\begin{align}
    \mathcal{L}_{tem}(G) = \frac{1}{N_l}\sum^{N-\Delta t}_{t=1}\sum^{t+\Delta t}_{t'=t+1}d(\mathbb{S}_{x}(t,t'), \mathbb{S}_{\hat{y}}(t,t')),
\end{align}
where $\Delta t$ is a hyper-parameter to adjust the length of training fragment, and we simply set it as $1$ for fast implementation.

\begin{figure}[!t]
    \centering
    \includegraphics[width=8cm]{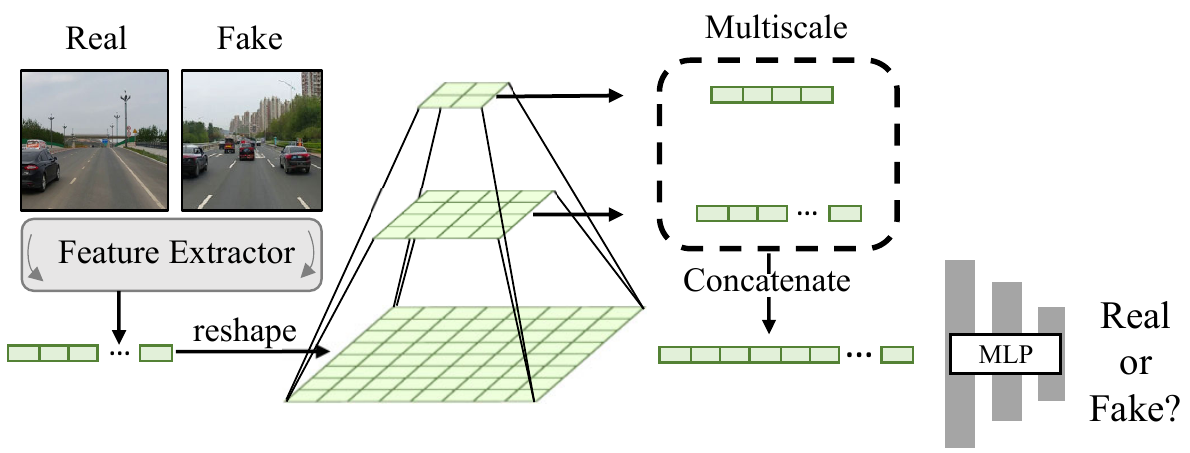}
    \caption{The framework of proposed multiscale region-wise discriminator. We obtain embedded tokens from \textit{Real} and \textit{Fake} frame via the feature extractor. Reshape tokens to make them spatially correspond to regions of the frames. Then we average spatially adjacent tokens on different sizes for different receptive fields and get the concatenated results. They are distinguished by the MLP module at last.}
    \label{fig:discriminator}
    \vspace{-0in}
\end{figure}

\subsection{Multiscale Region-Wise Discriminator}\label{method3.2}
In this paper, we design a multiscale region-wise discriminator to distinguish real or fake among visible references and synthesized target results, as shown in Fig.~\ref{fig:discriminator}. Firstly, we get the token embeddings $T_{y}, T_{\hat{y}} \in \mathbb{R}^{N_r \times d}$ from the pre-trained ViT and reshape them into the same format as the frame, then we average adjacent embeddings spatially with different sizes of receptive fields in non-overlapping manner. Taking size $2$ for an example, we average adjacent $4$ regions and get the collections $T_y^{2},T_{\hat{y}}^{2}\in \mathbb{R}^{\frac{N_r}{4}\times d}$ with twice larger receptive field than the origin. Similarly, we can also obtain $T_y^{4},T_{\hat{y}}^{4}\in \mathbb{R}^{\frac{N_r}{16}\times d}$ in a different scale, etc. By concatenating all new tokens within different scales, we have more informative representations, i.e., $\mathcal{T}_{y}$ = ($T_y^{2}$, $T_y^{4}$, ...) and $\mathcal{T}_{\hat{y}}$ = ($T_{\hat{y}}^{2}$, $T_{\hat{y}}^{4}$, ...). Then, a 3-layer learnable MLP network is used to classify whether the tokens are real or fake. Finally, the adversarial loss is detailed as:
\begin{align}
    \mathcal{L}_{adv}(G,D_r) &=\mathbb{E}_{y\sim p_{data}(y)}[\log D_r(y)] \nonumber\\
    &+\mathbb{E}_{x\sim p_{data}(x)}[\log (1-D_r(G(x)))],
\end{align}
where $D_r$ is our proposed multiscale region-wise discriminator, $D_r(y) = {\rm MLP}(\mathcal{T}_{y})$ and $D_r(G(x)) = {\rm MLP}(\mathcal{T}_{\hat{y}})$.

\subsection{Overall Optimization}
We train the network by minimizing the following losses:
\begin{align}
    \mathcal{L}_D =&-\mathbb{E}_{y\sim p_{data}(y)}[\log D_r(y)] \nonumber\\
    &-\mathbb{E}_{x\sim p_{data}(x)}[\log (1-D_r(G(x)))],\\
    \mathcal{L}_G =&~ \mathbb{E}_{x\sim p_{data}(x)}[\log (1-D_r(G(x)))]  \nonumber\\
    &+ \lambda_1 \cdot \mathcal{L}_{g}(G) + \lambda_2 \cdot \mathcal{L}_{l}(G) + \lambda_3 \cdot \mathcal{L}_{tem}(G),
\end{align}
where $\lambda_1$, $\lambda_2$ and $\lambda_3$ are tradeoff parameters.

\subsection{Discussion}

In nighttime infrared to daytime visible translation, the key challenge is to well preserve the structural details of infrared data. Perceptual loss \cite{PerceptualSimilarity, perceptual} and PatchNCE loss \cite{CUT} have been studied to constrain the structural consistency. Unfortunately, these losses do not explicitly decouple structural and style information. The quality of results is limited by different domain styles. 
Furthermore, F/LSeSim~\cite{FLSeSim} proposes a self-similarity strategy, which is enhanced by a small learnable network, to represent the domain-invariant structural information. However, the data augmentation manner utilized to train the small network is not suitable for reducing the stylistic effects of infrared data. We experimentally analyze the effect of style on similarity and visualize the cosine similarity between irrelevant regions in the Fig. \ref{fig:similarity}. Experiments indicate that the influence of infrared style is negative for self-similarity matching. On the contrary, our cross-similarity shows its ability of cutting off negative influences of style, which is tailored for representing the domain-invariant structural details from infrared data.

\section{InfraredCity Dataset}

\begin{figure}[!t]
  \centering
  \includegraphics[width=8.5cm]{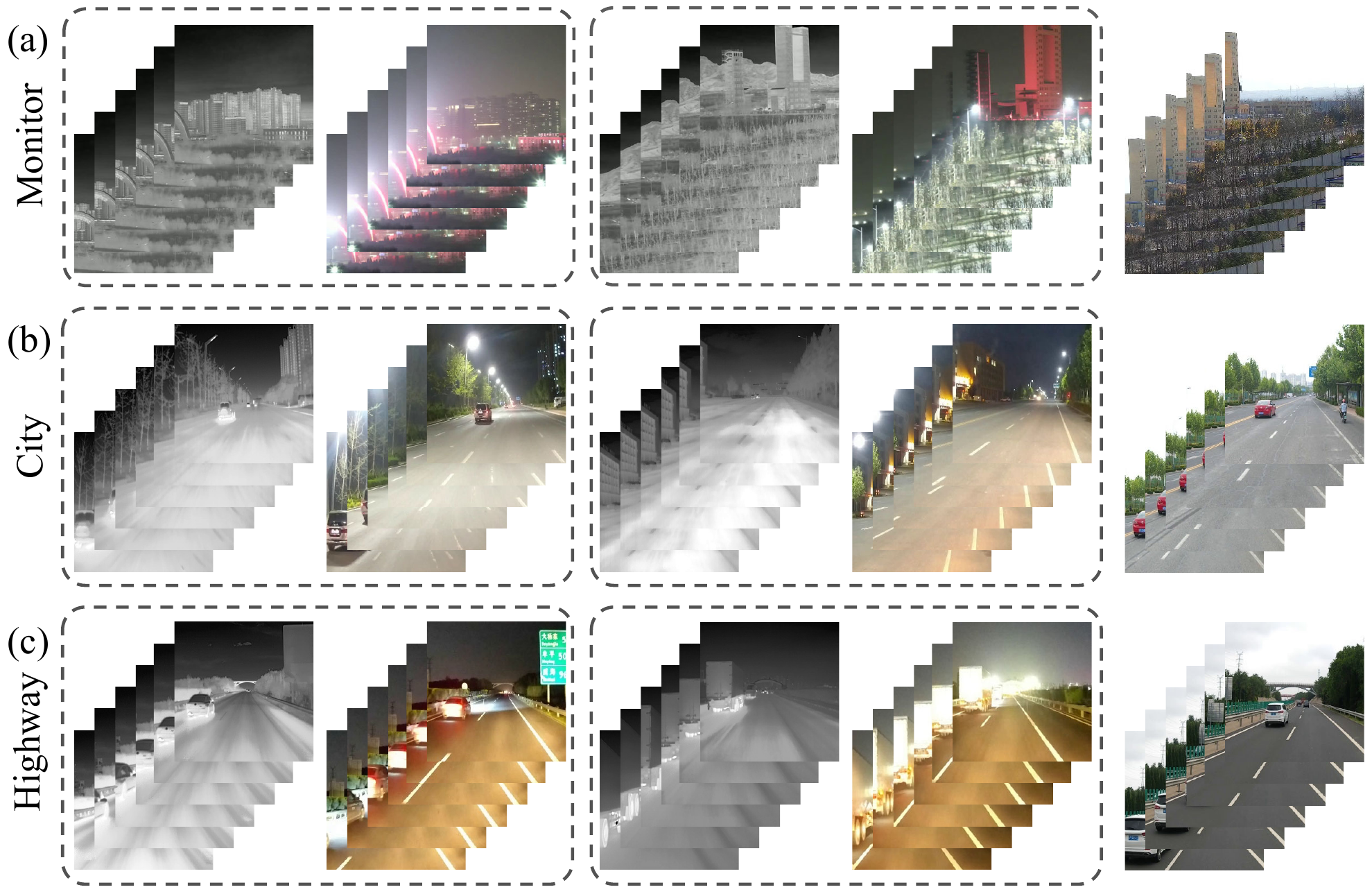}
  \caption{Several consecutive frames of InfraredCity-Lite in different scenes, (a) Monitor, (b) City and (c) Highway. All of them display two infrared and visible video clips captured at night with a clip of unambiguous daytime visible video. }
  \label{fig:Infred-Lite}
  \vspace{-0.15in}
\end{figure}

\begin{table}[!ht]
  \centering
  \small
  \caption{The structure of InfraredCity and InfraredCity-Lite}
  \label{tab:Dataset subsets.}
  \begin{tabular}{| m{10mm}<{\centering}| m{10mm}<{\centering} | m{10mm}<{\centering} | m{10mm}<{\centering} | m{10mm}<{\centering} | m{10mm}<{\centering} |}
      \hline
      \multicolumn{2}{|c|}{\textbf{InfraredCity}} & \multicolumn{4}{c|}{\textbf{Total Frame}} \\
      \hline
      \multicolumn{2}{|c|}{Nighttime Infrared} & \multicolumn{4}{c|}{201,856} \\
      \hline
      \multicolumn{2}{|c|}{Nighttime Visible} & \multicolumn{4}{c|}{201,856} \\
      \hline
      \multicolumn{2}{|c|}{Daytime Visible} & \multicolumn{4}{c|}{199,430} \\
      \hline\hline

      \multicolumn{2}{|c|}{\textbf{InfraredCity-Lite}} & \textbf{\makecell{Infrared\\Train}} &  \textbf{\makecell{Infrared\\Test}} & \textbf{\makecell{Visible\\Train}} & \textbf{Total} \\ 
      \hline

      \multirow{2}{*}{City} & clearday & 5,538 &1,000  & \multirow{2}{*}{5360}& \multirow{2}{*}{15,180}  \\
      \cline{2-4}
      & overcast & 2,282 & 1,000 &  & \\
      \cline{1-6}
      \multirow{2}{*}{Highway} & clearday & 4,412 & 1,000 &\multirow{2}{*}{6,463}  & \multirow{2}{*}{15,853} \\
      \cline{2-4}
      & overcast & 2,978 & 1,000 & & \\
      \hline
      \multicolumn{2}{|c|}{Monitor}  & 5,612 & 4,194 & 500 & 10,806\\
      \hline
  \end{tabular}
  \vspace{-0.15in}
\end{table}

Since a long-ware infrared camera conforms to the requirements of recognition and vehicle driving scenes at night, we utilize the binocular infrared color camera (DTC equipment) to capture nighttime infrared and visible videos. Additionally, a visible camera is adopted during the day to capture daytime visible videos in the same scenes, while having no sync with nighttime ones. It is well-known that deep models require massive amounts of training data. The lack of a high-quality relevant dataset limits the study of the translation from nighttime infrared to daytime visible. In such cases, our \textbf{InfraredCity} \footnote[1]{InfraredCity and InfraredCity-Lite will be released later.\label{InfraredCity}} dataset consists of \textbf{201,856} nighttime infrared frames, \textbf{201,856} nighttime visible frames, and \textbf{199,430} daytime visible frames, detailed in Tab.~\ref{tab:Dataset subsets.}. Our InfraredCity is about 20 times larger than the most relative dataset IRVI~\cite{I2V} which is provided for daytime infrared-to-visible video translation. Besides, we capture these infrared videos at night from three scenes (city, highway, and monitoring scenarios). Specifically, City and Highway are captured under clear and overcast weather conditions. The dataset is more challenging than the  since the domain gaps between nighttime infrared and daytime visible videos are much larger, raising requirements for preserving structural information of infrared videos covered by the gray appearance.

To facilitate comparison with other methods, we select parts of the InfraredCity dataset to build the \textbf{InfraredCity-Lite} \textsuperscript{\ref {InfraredCity}} dataset, which contains 41,839 frames in total. We design the InfraredCity-Lite for three forms: \textit{Single}, \textit{Double} and \textit{Triplet} to be in line with the input requirements of most image/video translation methods. The selection strategy is detailed in the supplementary material. Several examples of each scene are shown in Fig. \ref{fig:Infred-Lite}.

\begin{table}[!t]
  \centering
  \small
  \caption{Comparison of Infrared-Related Datasets}
  \vspace{-1em}
  \label{tab:Datasets comparision}
  \begin{tabular}{|m{28mm}<{\centering}|m{10mm}<{\centering}|m{7mm}<{\centering}|m{20mm}<{\centering}|}
      \hline
      \textbf{Name} & \textbf{Frame} & \textbf{Clip} & \textbf{Main Task} \\
      \hline
      InfraredCity & \textbf{603,142} & 9 & \multirow{3}{*}{video translation} \\
      \cline{1-3}
      InfraredCity-Lite & 41,839 & 9 &  \\
      \cline{1-3}
      IRVI \cite{I2V} & 24,352 & 12 &  \\
      \hline
      VOT2019 (RGBTIR) \cite{VOT2019} & 20,083 & 60 & object tracking \\
      \hline
      FLIR \cite{FLIR}& 4,224 & 1 & \multirow{2}{*}{object detection} \\
      \cline{1-3}
      KAIST (DAY ROAD) \cite{KAIST} & 16,176 & 9 & \\
      \hline
  \end{tabular}
  \vspace{-0.2in}
\end{table}

We comprehensively select four popular infrared related datasets (IRVI \cite{I2V}, VOT2019-RGBTIR \cite{VOT2019}, FLIR \cite{FLIR} and KAIST \cite{KAIST}) for comparison and details are shown in Tab. \ref{tab:Datasets comparision}. 
KAIST and FLIR are proposed for object detection and focus on capturing pedestrians on the street, which is not conducive to training an infrared-to-visible translation model with generalization. 
Although VOT2019-RGBTIR provides 60 infrared and visible video clips, its main challenge is object tracking in short video clips, which is inappropriate for generating agreeable visible results. IRVI consists of daytime infrared and daytime visible data under single weather, aligned at the hardware level. Although it is collected for infrared-to-visible translation tasks, it is more accessible for the generator to find a mapping compared with InfraredCity.

\section{Experiments}

\subsection{Datasets}
\textbf{InfraredCity-Lite} is collected for nighttime infrared to daytime visible translation. This dataset contains 37,339 training and 4500 testing frames. The resolution of both infrared and visible videos is $256 \times 256$. Our experiments are mainly based on it.

\textbf{IRVI} is a widely-popular dataset for infrared-to-visible translation, consisting of 22,080 training and 2272 testing frames. All of the videos are captured during the day, requiring models to affine the infrared video to visible results. The resolution is the same as the InfraredCity dataset.

\subsection{Experiment Setup}

\textbf{Evaluation Metrics.}
We first use the standard Fréchet Inception Distance (\textbf{FID}) \cite{FID} to compare the distribution of synthesized daytime visible frames with the distribution of real daytime visible ones from the features space. In the standard setting, these features are estimated by an InceptionV3 \cite{FID_network} pre-trained on the ImageNet \cite{ImageNet} dataset. The lower value of FID is better. Additionally, to further evaluate the improvement of translated results compared with nighttime visible, we report the \textbf{YOLO score} on object detection. YOLO score represents the quality of translated frames according to an off-the-shelf object detection network. In particular, we utilized the pre-trained YOLOv3 \cite{YOLOv3} to predict the vehicle occurred on scenes. By comparing the detection results against our manually annotated labels, we can acquire the YOLO scores with the standard detection metric Average Precision (AP). The higher YOLO scores indicate the more realistic the synthesized frames are.

\textbf{Implementation Details.}
We design our ROMA as a one-sided framework, consisting of a generator and a discriminator. Following \cite{cycleGAN}, we adopt the encoder-decoder architecture as our backbone network and apply this setting to all methods in our experiments for a fair comparison. The $L_{size}$ of our multiscale region-wise discriminator is set as $[3, 5, 7]$ for diverse receptive fields. For local structural consistency, the number of areas $N_a$ is $64$ and the size of areas is $75 \times 75$. For the balance of quality improvement and computation cost, $\varDelta t$ is set as 2 for $\mathcal{L}_{tem}$. For the constraints of ROMA, the hyperparameter $\lambda_1$, $\lambda_2$ and $\lambda_3$ are set as $5.0$, $5.0$ and $1.0$.

\begin{table}[!tb]\scriptsize
  \centering
  \caption{\small Fréchet Inception Distance (FID) of different translation methods on InfraredCity-Lite. Lower is better.}
  \setlength{\tabcolsep}{2.8pt}
  \def\arraystretch{1.2}
  \label{tab:InfredFID}

  \begin{tabular}{|m{16mm}<{\centering}|m{6mm}<{\centering}  m{6mm}<{\centering}  m{6mm}<{\centering} | m{6mm}<{\centering} m{6mm}<{\centering} m{6mm}<{\centering} | m{6mm}<{\centering} | m{8mm}<{\centering}|}
  \hline
  \multirow{4}{*}{\textbf{Method}} & \multicolumn{8}{c|}{\textbf{FID}} \\
  \cline{2-9}
   & \multicolumn{7}{c|}{Traffic} & \multirow{3}{*}{Monitor}  \\
  \cline{2-8}
   & \multicolumn{3}{c|}{City} &\multicolumn{3}{c|}{Highway} & \multirow{2}{*}{all} &\\
   \cline{2-7}
   & clear & overcast & all & clear & overcast & all & & \\
  \hline
  CUT & 0.5809 & 0.5607 & 0.6086 & 0.4544 & 0.5133  & 0.4739 & 0.4089 & 0.9785 \\
  CycleGAN& 0.6299 & 0.5879 & 0.7125 & 0.4787 &  0.5489 & 0.4920 & 0.4204 & 0.8129 \\
  F/LSeSim& 0.4984 & 0.5369 & 0.4834 & 0.5108 & 0.5288  & 0.4809 & 0.2724 & 0.8984 \\
  RecycleGAN& 0.5942 & 0.5974 & 0.5969 & 0.5173 &  0.5998 & 0.5101 & 0.3431 & 0.9433 \\
  MoCycleGAN & 0.5117 & 0.5346 & 0.5011 & 0.5029 &  0.5976 & 0.4791 & 0.3163 & 0.7298\\
  I2V-GAN & 0.5052 & 0.5574 & 0.4649 & 0.5064 & 0.5105 & 0.4515  & 0.2872 & 0.7039 \\
  \hline
  Ours w/o $D_r$  &0.4765 & 0.5358 & 0.4594 & 0.4623 & 0.4926 & 0.4503  & 0.2653 & 0.7322 \\
  Ours w/o $\mathcal{L}_{g}$ & 0.4601 & 0.5296 & 0.4028 & 0.4133 & 0.3970 & 0.3924  & 0.2493 & 0.6015 \\
  Ours w/o $\mathcal{L}_{l}$ &0.4222 & 0.5244 & 0.4117 & 0.3935 & 0.4081 & 0.3922  & 0.2174 & 0.6913 \\
  Ours w/o $\mathcal{L}_{tem}$ & 0.4295 & 0.5013 & 0.4699 & 0.3872 & 0.4583 & 0.4155 & 0.2251  & 0.5751   \\
  Ours & \textbf{0.4018} & \textbf{0.5149} & \textbf{0.3929} & \textbf{0.3325} & \textbf{0.3823} & \textbf{0.3444}  & \textbf{0.2002} & \textbf{0.5488} \\
  \hline
  \end{tabular}
  \vspace{-0.2in}
\end{table}

\begin{table}[!tb]\scriptsize
  \centering
  \caption{\small Fréchet Inception Distance (FID) of different translation methods on IRVI. Lower is better.}
  \setlength{\tabcolsep}{2.8pt}
  \def\arraystretch{1.2}
  \vspace{-0.15in}
  \label{tab:IRVI}
  \begin{tabular}{|m{16mm}<{\centering} | m{9mm}<{\centering} | m{7mm}<{\centering} m{7mm}<{\centering} m{7mm}<{\centering} m{7mm}<{\centering} m{7mm}<{\centering} m{8mm}<{\centering}|}
  \hline
  \multirow{3}{*}{\textbf{Method}} & \multicolumn{7}{c|}{\textbf{FID}} \\
  \cline{2-8}
    & \multirow{2}{*}{Traffic} & \multicolumn{6}{c|}{Monitoring} \\
  \cline{3-8}
   & & sub-1 & sub-2 & sub-3 & sub-4 & sub-5 & all \\
  \hline
  CUT &  0.5739 & 1.4348 & 0.6731  & 2.2294  & 1.0553 & 2.1973 & 1.0893 \\
  CycleGAN &  0.6714 & 1.4027 & 0.8056 & 2.1497 & 1.0359 & 1.6266 & 0.8792 \\
  F/LSeSim &  0.4321 & 1.5135 & 0.5452 & 1.6665 & 0.9457 & 1.9594 & 0.9232 \\
  MoCycleGAN &  0.7911 & 1.5556 & 0.9847 & 2.5013 & 1.1040 & 2.1171 & 1.0515\\
  RecycleGAN &  0.5255 & 1.6680 & 0.7521 & 2.0387 & 1.2959 & 1.8518 & 1.0609 \\
  I2V-GAN &  0.4425 & 1.4840 & 0.5905 & 1.7916 & 0.9189 &  1.6015 & 0.8715 \\

  \hline
  Ours w/o $D_r$ & 0.4061 & 1.3427 & 0.3383 & 1.6095 & 0.6155 & 1.5241 & 0.9036\\
  Ours w/o $\mathcal{L}_g$ & 0.3873 & 1.3367 & 0.3756 & 1.6163 & 0.6084 & 1.5830 & 0.7935 \\
  Ours w/o $\mathcal{L}_l$ & 0.3824 & 1.2759 & 0.2723 & 1.5064 & 0.5453 & 1.5402 & 0.7437\\
  Ours w/o $\mathcal{L}_{tem}$ & 0.3633 & 1.2527 & 0.2849 & 1.5034 & 0.5518 & 1.4733 & 0.7465 \\
  Ours  & \textbf{0.3467} & \textbf{1.2301} & \textbf{0.2485} & \textbf{1.4765} & \textbf{0.5188} & \textbf{ 1.4438} & \textbf{0.7334} \\
  \hline
  \end{tabular}
  \vspace{-0.25in}
\end{table}

\begin{figure*}[ht]
  \centering
  \includegraphics[width=17.3cm]{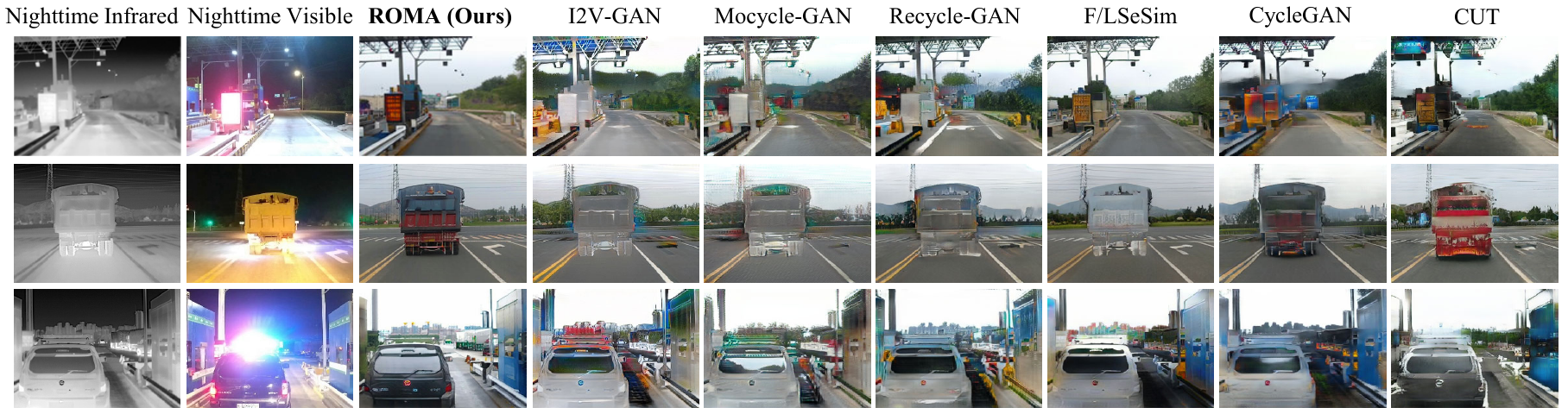}
  \caption{Qualitative comparisons of different methods in InfraredCity-Lite. Our ROMA has obvious advantages in detail.}
  \label{fig:InfraredCity_compare}
  \vspace{-0.2in}
\end{figure*}

\begin{figure}[t]
  \centering
  \includegraphics[width=8.3cm]{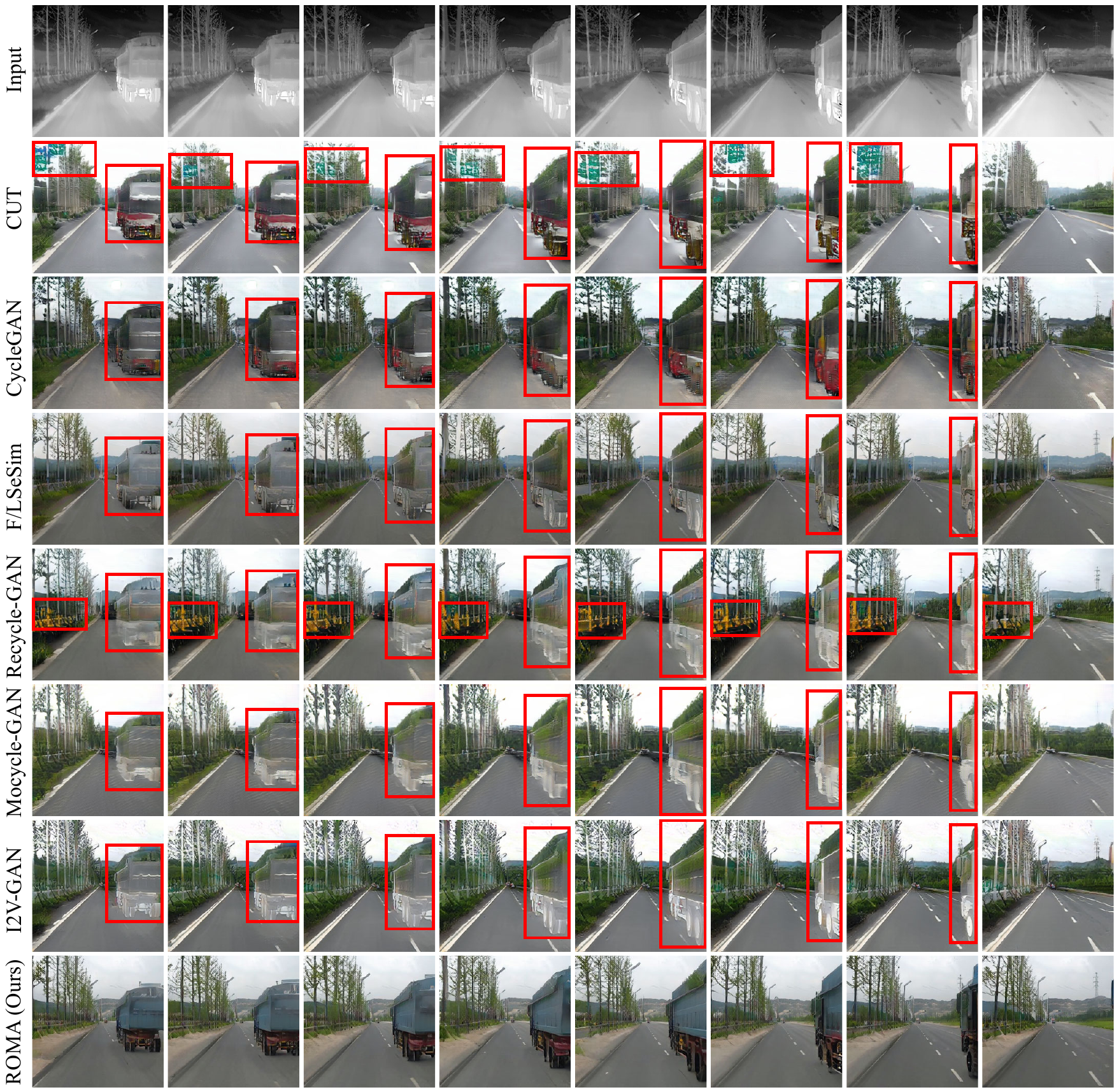}
  \vspace{-1em}
  \caption{Translation results of Highway. ROMA takes advantage of preserving structural and temporal consistency.}
  \label{fig:video_compare}
  \vspace{-0.15in}
\end{figure}

\begin{figure}[t]
  \centering
  \includegraphics[width=8.3cm]{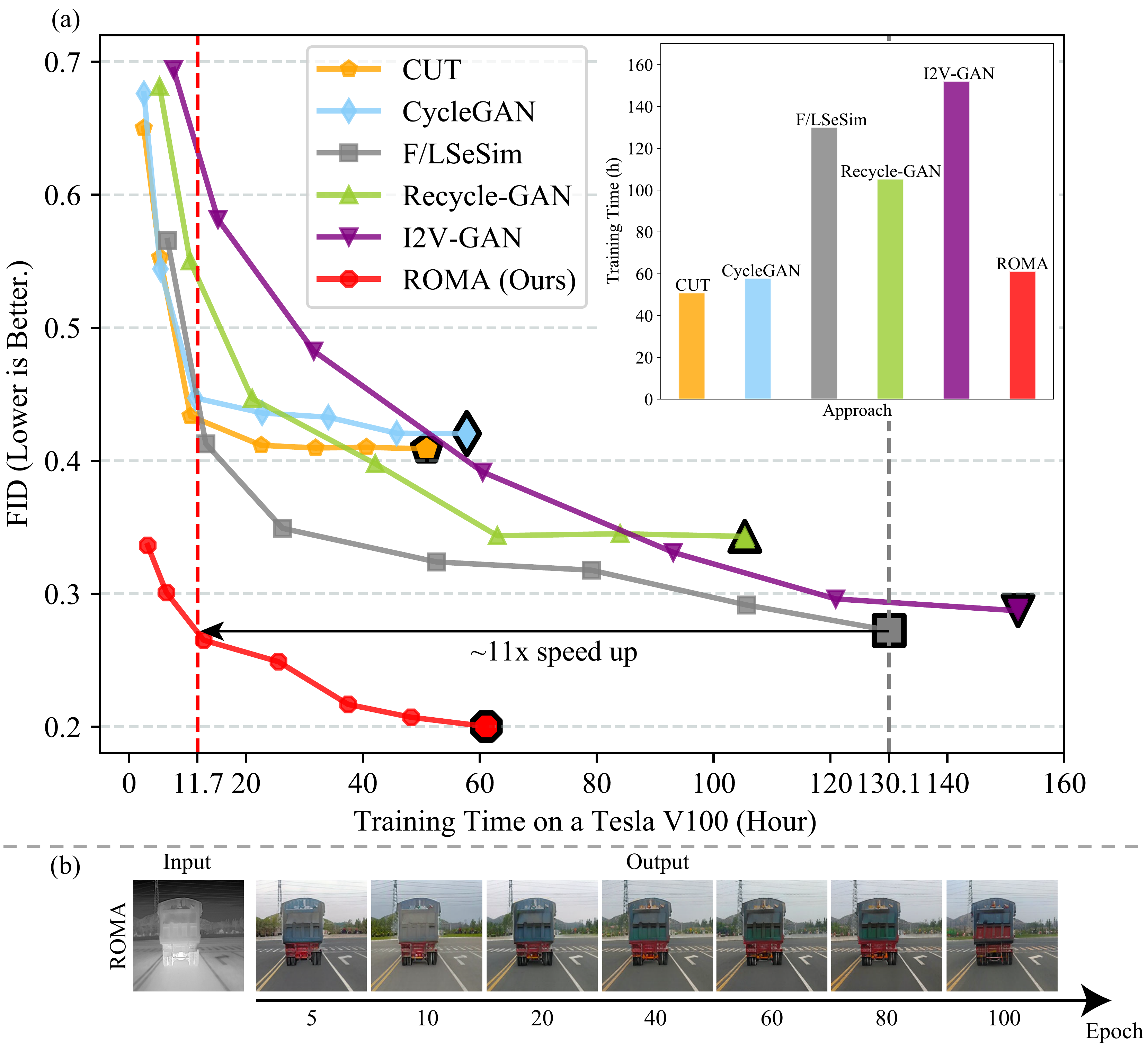}
  \caption{(a) Line chart of FID at different time nodes. We run all methods for 100 epochs while maintaining the same amount of training data for each epoch. Nodes with black borders represent the end time. (b) We show the quality of the ROMA results at different moments, corresponding to the moments of ROMA in (a).}
  \label{fig:time_compare}
  \vspace{-0.2in}
\end{figure}

\subsection{Comparisons with Other Methods}

We compare our method with several state-of-the-art methods of unpaired image-to-image and video-to-video translation, i.e., CUT \cite{CUT}, CycleGAN \cite{cycleGAN}, F/LSeSim \cite{FLSeSim}, Recycle-GAN \cite{RecycleGAN}, Mocycle-GAN \cite{MocycleGAN} and I2V-GAN \cite{I2V} in terms of FID results, ablation study, qualitative comparisons, and training time.

\textbf{Results of InfraredCity-Lite.}
As shown in Tab. \ref{tab:InfredFID}, our ROMA obtains state-of-the-art performance, which indicates that ROMA can substantially improve the quality of generated daytime visible frames. 
Especially, F/LSeSim which only focuses on maintaining structure achieves similar results with the tailored infrared-to-visible translation method, i.e., I2V-GAN. It confirms the key challenge of InfraredCity-Lite is to maintain the structural information of infrared videos. Moreover, compared with them, our ROMA outperforms them by 22.3\% on average relatively on the FID metric. Especially, 35.5\% improved relatively on the clear Highway scene. These improvements indicate the advantages of ROMA for generating excellent daytime visible results, especially for preserving domain-invariant structure.

\textbf{Resuls of IRVI.} 
Our main evaluation is thus against the state-of-art I2V-GAN on IRVI. The results of the comparison are presented on Tab. \ref{tab:IRVI}. Notably, our ROMA achieves the top-ranked FID performance on all scenes and outperforms I2V-GAN 26.2\% on average relatively, surpassing by 57.9\% relatively on the sub-2 scene. This again validates our effectiveness in maintaining infrared structure with cross-similarity.

\textbf{Ablation Study.} 
We conduct ablation experiments on both two datasets to study how each design of ROMA influences the whole framework and the results are shown in Tab. \ref{tab:InfredFID} and Tab. \ref{tab:IRVI}. Our region-wise cross-similarity matching constraints ($\mathcal{L}_g$, $\mathcal{L}_l$ and $\mathcal{L}_{tem}$) have great contributions to improving the quality of generated visible results. Notably, the improvement of multiscale region-wise discriminator is not inferior to cross-similarity.

\textbf{Qualitative Comparisons.}
Furthermore, we make a qualitative comparison with other approaches on the InfraredCity-Lite dataset. From Fig. \ref{fig:InfraredCity_compare} and Fig. \ref{fig:video_compare}, it is observed that the synthesized daytime visible frames by ROMA have better visual performance compared with other state-of-the-art methods. Notably, our generated frames keep a better correspondence on structure with the input infrared frames. This benefits from our domain-invariant representation, cross-domain region similarity. The constraints of other approaches help the generator learn the structural information comfortably at the beginning but heavily as the training process proceeds due to the prominent difference between source and target styles. In contrast, our cross-similarity matching technique gets rid of the negative influence of style all the time, which helps the generator to learn structural information consistently and comfortably.

Furthermore, as depicted in Fig. \ref{fig:video_compare}, the baselines all generate visible frames whose overall color is somewhat inaccurate and especially disappoint us on the generated truck. In contrast, our ROMA demonstrates pleasant results which are more analogous to the style of unambiguous daytime visible videos. Besides, favorable temporal coherence in our results can be observed in these results.
We display more vivid synthesized videos on the supplementary material. All of them confirm the capability of our temporal cross-similarity consistency.

\textbf{Training Time.}
Besides, we make a comparison on efficiency since it is critical for applications in reality. Fig. \ref{fig:time_compare} (a) showcases the efficiency of the different methods under a unified standard. Our ROMA achieves the best performance compared with other approaches. Especially, to obtain the best FID scores of other approaches, only approximately 11.7 hours are cost by ROMA, which is about 11 times faster than F/LSeSim. Besides, I2V-GAN consists of many hand-designed constraints from feature level and pixel level for spatial and temporal consistency. These bring improvements yet arduous convergence. On the contrary, we concentrate on the fundamental information of infrared videos, structural information covered by the gray appearance. Thus, the designed domain-invariant representation, cross-similarity not only maintains structure well but also helps the generator learn the structural information candidly. The results of Fig. \ref{fig:time_compare} (b) make a further demonstration of the efficiency of our ROMA. It indicates that ROMA gains excellent details even in the early stage of training.

\begin{figure}[!t]
  \centering
  \includegraphics[width=8.3cm]{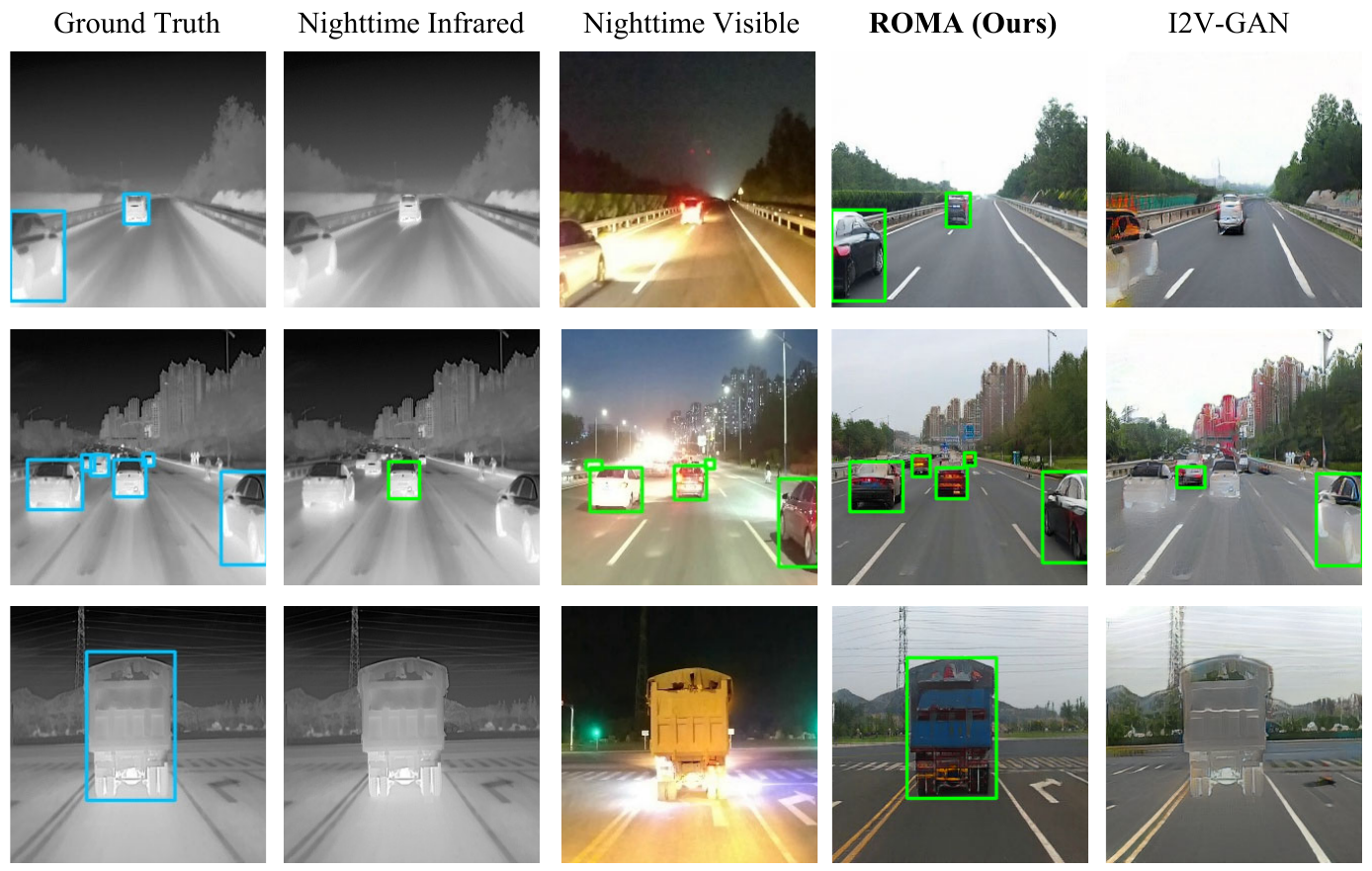}
  \caption{Comparisons of vehicle detection results on nighttime infrared, nighttime visible and translated daytime visible. Detection tasks are performed by the pre-trained YOLOv3 \cite{YOLOv3} model. Our goal is to generate the translated results as similar as possible to actual visible ones, especially in detail.}
  \label{fig:detection_compare}
  \vspace{-0.15in}
\end{figure}

\begin{table}[!ht]
  \centering
  \small
  \caption{Comparison of YOLO scores (\%) for vehicle detection.}
  \label{tab:detection_compare}
  \begin{tabular}{ m{8mm}<{\centering} | m{13mm}<{\centering}  m{13mm}<{\centering}  m{13mm}<{\centering}  m{13mm}<{\centering} }
    \toprule
      \textbf{Scenes} & \textbf{Nighttime Infrared} & \textbf{Nighttime Visible} & \textbf{I2V-GAN} & \textbf{ROMA (Ours)} \\
      \midrule
      \textbf{AP} & 25.0 & 26.1  & 32.2& \textbf{50.1}\\
      \bottomrule

  \end{tabular}
  \vspace{-0.1in}
\end{table}

\begin{figure}[!t]
  \centering
  \includegraphics[width=8.3cm]{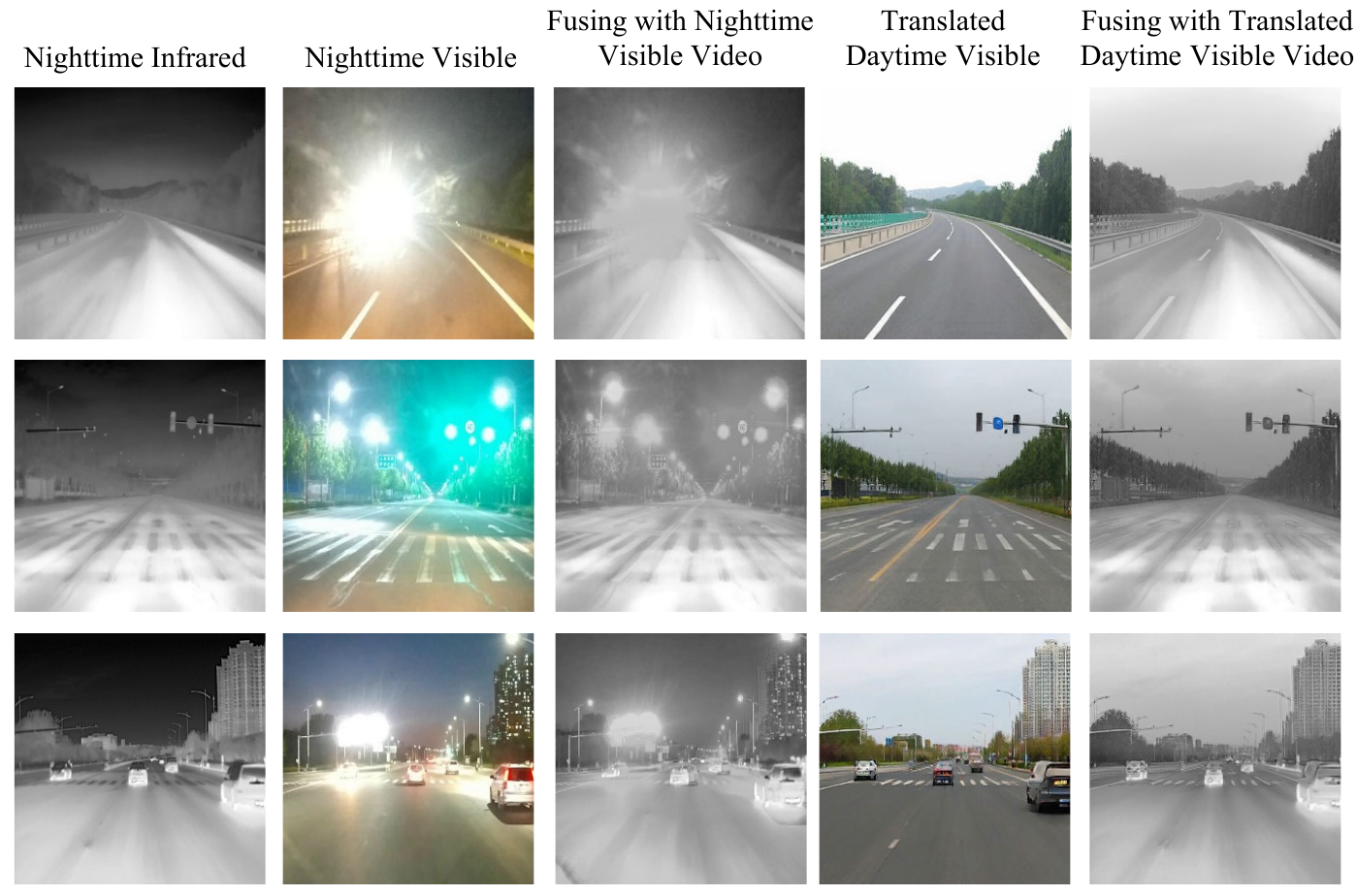}
  \caption{Comparisons of video fusion results. Fusing nighttime infrared videos with awful nighttime visible videos brings blurry results while the fusion with translated daytime visible videos contributes to unambiguous results. }
  \label{fig:fusion_compare}
  \vspace{-0.2in}
\end{figure}

\subsection{Object Detection}

Object detection is a fundamental and core problem in computer vision. Studies \cite{yolo, SSD, YOLOv3} achieve remarkable improvements with the advantages of large-scale annotated datasets. However, the object detection model is often vulnerable to data variance, especially when operating at night. In such cases, the translation from stable nighttime infrared to daytime visible is an ideals solution.

For further evaluating the quality of generated results, we utilize a pre-trained YOLOv3 \cite{YOLOv3} model for object detection (e. g., vehicle detection) on nighttime infrared, nighttime visible, and translated daytime visible videos. From Fig. \ref{fig:detection_compare}, we can observe that the translated daytime videos indeed favor the detection. Besides, the improvements of our ROMA are detailed on Tab. \ref{tab:detection_compare} via YOLO scores. More comparisons are displayed in the supplementary material.

\subsection{Video Fusion}

Generally, the infrared (IR) vision technique is adopted for the context enhancement in nighttime vision by fusing it with the visible (VI) image \cite{zhou2016fusion, 2018fusion,lowRankFusion}. However, IR/VI image fusion methods are only ideals at dawn when the visible camera can still capture the relatively clear visible scene. 
Additionally, it is challenging for fusing infrared videos with visible ones since they are not paired at the pixel level and videos contain the nature of temporal coherence. 

On the contrary, the translated daytime visible results from ROMA are unambiguous and match the infrared input videos at the pixel level. Observed from Fig. \ref{fig:fusion_compare}, the fusion results of nighttime infrared frames and translated daytime visible frames are more distinct and semantic compared with the ones of nighttime infrared and nighttime visible frames.
We display more fusion results in the supplementary material.

\section{Conclusion}

In this paper, we introduce a tailored framework ROMA to translate the unpaired nighttime infrared videos into fine-grained daytime visible ones via the proposed cross-domain region similarity matching technique, which effectively transferring the structural knowledge of the infrared data and preserving the spatiotemporal consistency. To further enhance the reality of translated videos, a multiscale region-wise discriminator is designed, and extensive experiments validate that ROMA obtains state-of-the-art performance when producing unambiguous daytime visible videos. Moreover, tests on nighttime object detection and video fusion tasks demonstrate that ROMA can generate reliable results for night vision applications. Besides, we provide a challenging dataset for nighttime infrared to daytime visible video translation, i.e., \textit{InfraredCity}, and we hope this will encourage more researches in this area.

% \clearpage
\bibliographystyle{ACM-Reference-Format}
\bibliography{reference}

\clearpage

\section*{SUPPLEMENTARY}
\section{Datasets}

It is well known that training deep models relies on massive amounts of training data. Thus, a large-scale dataset is required for the unpaired video translation between nighttime infrared and daytime visible. Unfortunately, existing infrared-related datasets are either collected for other specific research areas or limited by the small scale. 
To this end, we capture the infrared and visible videos to build the \textit{InfraredCity} dataset, by far the largest infrared-related dataset to date. Furthermore, we artificially select parts of the InfraredCity dataset to make the \textit{InfraredCity-Lite} to facilitate comparison with other methods.

\subsection{InfraredCity}

The infraredCity dataset consists of nighttime infrared, nighttime visible, and daytime visible videos. In particular, the nighttime infrared and nighttime visible videos are captured through a binocular infrared color camera (DTC 300 equipment), which is aligned at the hardware level. 

\paragraph{Nighttime Infrared and Nighttime Visible Videos.}
According to the heat-related imaging principle, infrared sensors can be divided into the near-infrared camera and the long-wave infrared camera. Compared with the near-infrared camera, the long-ware infrared camera conforms to the requirements of recognition and vehicle driving scenes at night. Thus, we utilize a type of long-wave equipment, DTC 300 for capturing infrared videos. 
As for the nighttime visible videos, they are still captured via the DTC 300 equipment. 
Since we perform the alignment of the infrared and the visible cameras at the hardware level, nighttime infrared and nighttime visible videos are identical in scene and length.

For practical applications at night, we shoot traffic and monitoring videos to build the InfraredCity dataset. Traffic videos consist of scenes from the city and highway. 
The more diverse the content of city videos is, the harder it will be for the model to converge.
Thus, the city scenes captured on the InfraredCity focus on varied things (e.g., buildings and cars) which could be observed in our daily life when driving. On the other hand, motion shift should be one of the challenges in video-related datasets. Thus, we capture highway scenes for rapid movement changes. Additionally, since infrared sensors are widely used for monitoring, we also collect the monitoring videos as a part of InfraredCity. Instead of fixing the angle, we rotate the camera in all directions for diversity and motion shift.

\paragraph*{Daytime Visible Videos.}
Since we want high-quality translated daytime visible results, we shoot the daytime visible videos on a clear day. The daytime visible videos still contain scenes of the city, highway, and monitoring. Although all scenes in the daytime visible videos correspond to the scenes in the nighttime infrared videos, there are large differences in each scene. For example, trucks are more common at night than during the day because there are fewer vehicles on the nighttime road, which benefits the trucks. The inconsistency of the shooting time between nighttime infrared videos and daytime visible ones brings great challenges to the dataset.

\subsection{InfraredCity-Lite}

Although the InfraredCity is diverse and large-scale, it may not be appropriate for experimental studies due to repetition of similar frames and mixture of all scenes. In this case, we select parts of InfraredCity to build the InfraredCity-Lite. 

\paragraph*{Scenario.}
To be in line with the input requirements of most image/video translation methods, three scenarios: \textit{Single}, \textit{Double} and \textit{Triplet} are designed on InfraredCity-Lite. In particular, The ratio of the three is controlled at about $3:2:1$. Taking Triplet as an example, we select three consecutive frames from ten consecutive frames to form the Monitor part of the InfraredCity-Lite dataset. The strategy reduces the repetition of similar frames on our dataset. It is more suitable in size and structure as an experimental dataset for research.

\paragraph*{Structure.}
Unlike mixing all scenes on InfraredCity, InfraredCity-Lite is divided into \textit{City}, \textit{Highway}, and \textit{Monitor}, which correspond to city scenes, highway scenes, and monitoring scenes on InfraredCity. Notably, nighttime infrared videos on City and Highway are captured under two weather conditions, \textit{clearday} and \textit{overcast}. The clear day condition is ideal for infrared sensors while the overcast condition may affect the temperature contrast. Thus, taking overcast weather into account can help InfraredCity-Lite to be more applicable for the actual scene. Additionally, since the main challenges of monitoring scenes are the different angles of the monitor, we rotate the monitor in all directions and collected videos named Monitor. The structure of InfraredCity-Lite can be viewed on paper.

\begin{figure*}[ht]
  \centering
  \includegraphics[width=16.3cm]{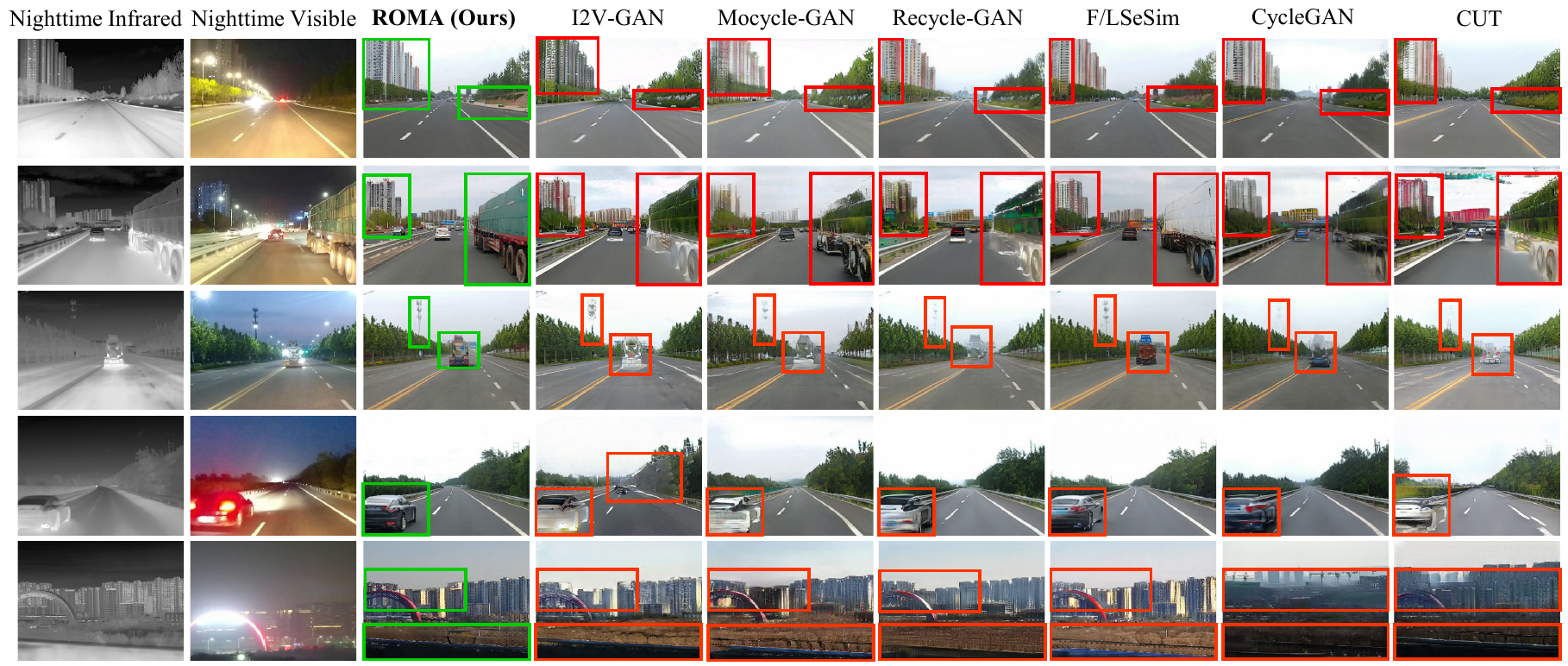}
  \vspace{-1em}
  \caption{Qualitative comparisons of different methods in InfraredCity-Lite. Our ROMA has obvious advantages in detail.}
  \label{fig:local_global}
  \vspace{-0.1in}
\end{figure*}

\begin{figure}[ht]
  \centering
  \includegraphics[width=8.3cm]{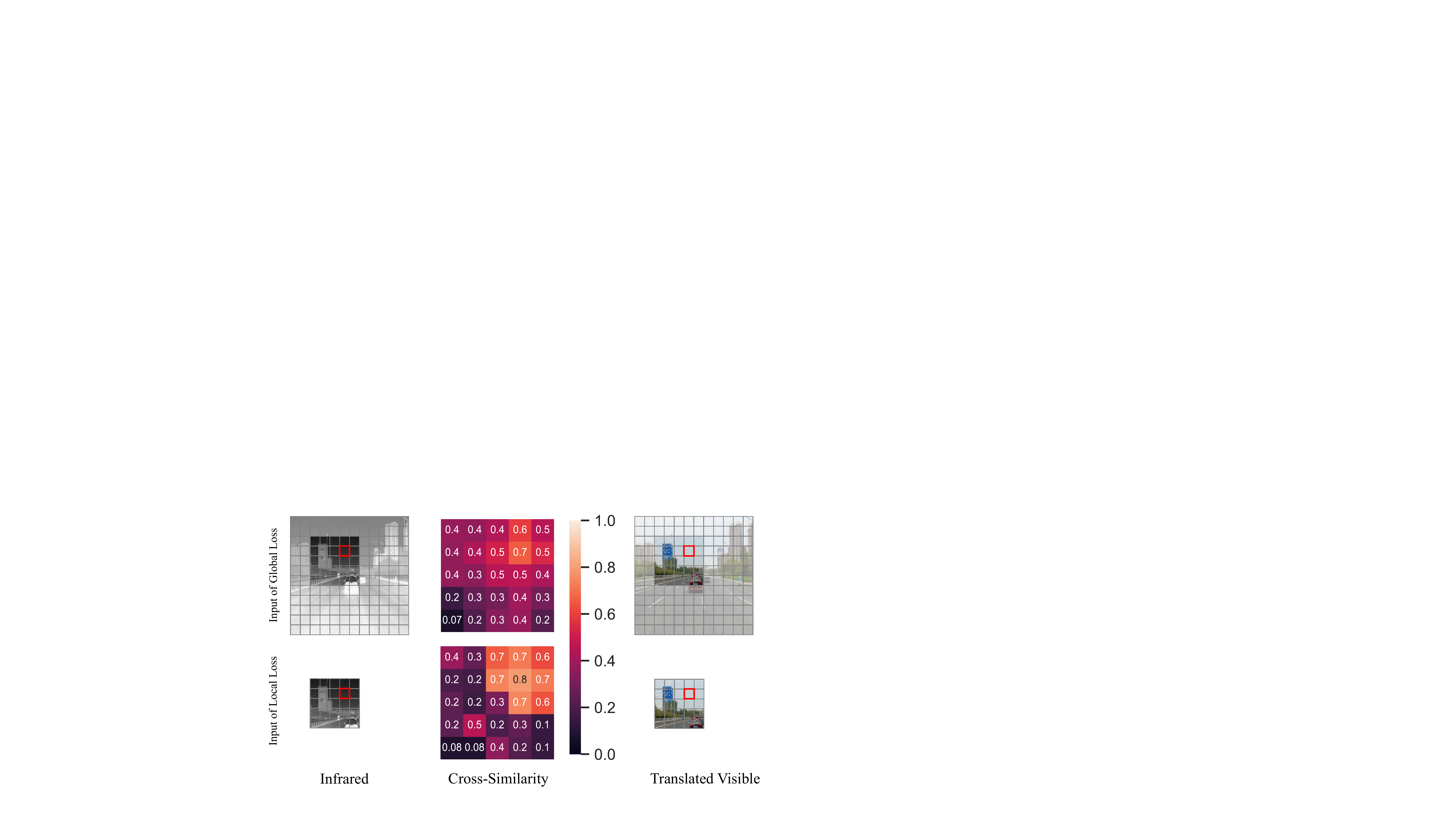}
  \vspace{-1em}
  \caption{Visualization of difference between $\mathcal{L}_g$ and $\mathcal{L}_l$. It is obvious that $\mathcal{L}_l$ is more sensitive in local details compared with $\mathcal{L}_g$.}
  \label{fig:image_comparison}
  \vspace{-0.15in}
\end{figure}

\begin{table}[!t]
  \centering
  \scriptsize
  \caption{Comparison of Infrared-Related Datasets}
  \vspace{-1em}
  \renewcommand{\arraystretch}{1.5} % Default value: 1
  \label{tab: comparision}
  \begin{tabular}{c|ccccc}
    \toprule
    \textbf{Dataset} &\textbf{Frame} & \textbf{Clip} & \textbf{Type} &\textbf{Clean} & \textbf{Task}  \\
    \hline
    InfraredCity & 603,142 & 9 & Video & No & Video Translation  \\
    \hline
    InfraredCity-Lite & 41,839 & 9 & Video\&Image & Yes & Video Translation   \\
    \hline
    IRVI & 24,352 & 12 & Video & No & Video Translation   \\
    \hline
    VOT2019 (RGBTIR) & 20,083 & 60 & Video & Yes & Object Tracking   \\
    \hline
    FLIR & 4,224 & 1 & Image & Yes & Object Detection   \\
    \hline
    KAIST (DATROAD) & 16,176 & 9 & Video & Yes & Object Detection   \\
    \bottomrule
  \end{tabular}
  \vspace{-0.15in}
\end{table}

\subsection{Other Infrared-Related Datasets}

We compare InfraredCity and InfraredCity-Lite with other datasets (IRVI \cite{I2V}, VOT2019-RGBTIR \cite{VOT2019}, FLIR \cite{FLIR} and KAIST \cite{KAIST}) and summarize the differences in Tab. \ref{tab: comparision}. 

\paragraph*{Frame.} 
It is well-known that deep models require massive amounts of training data. Our InfraredCity contains 603,142 video frames in total, which is about twenty times larger than the recently released IRVI (24,352) in number. We can observe that even the proposed InfraredCity-Lite dataset is nearly double larger than the other four datasets on average.

\paragraph*{Main Task.}
The IRVI dataset is released for unpaired daytime infrared to daytime visible translation. As for VOT2019 (RGBTIR), it is utilized for object tracking. Besides, both FLIR and KAIST (DAY ROAD) are used to evaluate the ability of the model on object detection. On the contrary, our InfraredCity and InfraredCity-Lite are collected for the unpaired video translation between nighttime infrared videos and daytime visible ones.

\paragraph*{Others.} VOT2019 (RGBTIR), FLIR, KAIST (DAY ROAD), and our InfraredCity-Lite are manually cleaned before being released. Additionally, FLIR is a small dataset and utilized for object detection at the image level.

\section{More Experiments}

\paragraph*{Difference between $\mathcal{L}_g$ and $\mathcal{L}_l$.} We further analyze the differences between \textit{global cross-similarity for structural consistency} and \textit{local cross-similarity for structural consistency} and visualize them in Fig. \ref{fig:local_global}. Although the cross-similarity under the global strategy can distinguish different regions, the local strategy brings more distinct cross-similarity results, which indeed favors the model to learn more details in the local area. In a word, both $\mathcal{L}_g$ and $\mathcal{L}_l$ optimize the generator globally and locally, respectively.

\paragraph*{Quality comparisons.} We additionally display comparisons of different methods on Fig. \ref{fig:image_comparison}. Our ROMA outperforms the others with better details (e.g., cars and buildings). More displays of results (e.g., translated videos, object detection and video fusion) will be released later. Our ROMA can achieves well-perserved content structures and desirable style patterns.

\begin{table}[!ht]
  \centering
  \small
  \caption{The comparison of subjective evaluations. The score is from 1 to 10 and the higher score is better.}
  \label{tab:userstudy}
  \begin{tabular}{ m{23mm}<{\centering} | m{12mm}<{\centering}  m{12mm}<{\centering} m{18mm}<{\centering} }
    \toprule
      \textbf{Method} & \textbf{Realism} & \textbf{Fluency}  & \textbf{Improvement}\\
      \midrule
      ROMA (Ours) & \textbf{8.5} & \textbf{9.0} & \textbf{6.0}\\
      \midrule
      I2V-GAN \cite{I2V} & 6.2 & 8.1 & 2.2\\
      \midrule
      Mocycle-GAN \cite{MocycleGAN} & 6.0 & 7.6 & 1.8 \\
      \midrule
      Recycle-GAN \cite{RecycleGAN} & 5.2 & 6.8 & 1.1 \\
      \bottomrule
  \end{tabular}
  \vspace{-0.1in}
\end{table}

\paragraph*{User study.} We conduct an additional user study for the comparison of subjective evaluations on Tab. \ref{tab:userstudy}, which are assessed by 25 professional researchers. We select several translated videos on InfraredCity from different unpaired video-related methods (I2V-GAN \cite{I2V}, Mocycle-GAN \cite{MocycleGAN} and Recycle-GAN \cite{RecycleGAN}). Realism and Fluency are generally utilized for evaluating videos. Moreover, the improvement score is used to measure how much additional information the user can obtain from the translated daytime visible results compared to the nighttime visible videos. Our ROMA outperforms the others in the three metrics.

\end{document}